\pdfoutput=1

\documentclass[11pt]{article}

\usepackage{acl}

\usepackage{times}
\usepackage{latexsym}

\usepackage{amssymb}
\usepackage{graphicx}
\usepackage{subcaption}
\usepackage{tabularx} 
\usepackage{mathptmx}
\usepackage{amsmath, calc}

\usepackage{ctable}

\newcommand{\x}{\boldsymbol{x}}
\newcommand{\y}{\boldsymbol{y}}

\usepackage[T1]{fontenc}

\usepackage[utf8]{inputenc}

\usepackage{microtype}

\title{Efficiently Aligned Cross-Lingual Transfer Learning for Conversational Tasks using Prompt-Tuning}

\author{Lifu Tu\Thanks{Equal contribution}, Jin Qu$^*$ \\ \textbf{Semih Yavuz, Shafiq Joty, Wenhao Liu \thanks{~~Work was done when the author was a full time employee at Salesforce Research}, Caiming Xiong, Yingbo Zhou}  \\
  Salesforce AI Research \\
  \texttt{\{ltu,jqu,syavuz,sjoty,cxiong,yingbo.zhou\}@saleforce.com} 
  \\}

\begin{document}
\maketitle
\begin{abstract}

Cross-lingual transfer of language models trained on high-resource languages like English has been widely studied for many NLP tasks, but focus on conversational tasks has been rather limited.
This is partly due to the high cost of obtaining non-English conversational data, which results in limited coverage. In this work, we introduce XSGD\footnote{\url{https://console.cloud.google.com/storage/browser/multilingual-sgd-data-research}} for cross-lingual alignment pretraining,  a parallel and large-scale multilingual conversation dataset that we created by translating the English-only Schema-Guided Dialogue (SGD) dataset~\citep{rastogi2020towards} into 105 other languages. XSGD contains about 330k utterances per language. To facilitate aligned cross-lingual representations, we develop an efficient prompt-tuning-based method for learning alignment prompts. We also investigate two different classifiers: NLI-based and vanilla classifiers, and test cross-lingual capability enabled by the aligned prompts. We evaluate our model's cross-lingual generalization capabilities on two conversation tasks: slot-filling and intent classification. Our results demonstrate strong and efficient modeling ability of NLI-based classifiers and the large cross-lingual transfer improvements achieved by our aligned prompts, particularly in few-shot settings.
We also conduct studies on large language models (LLMs) such as text-davinci-003 and ChatGPT in both zero- and few-shot settings. While LLMs exhibit impressive performance in English, their cross-lingual capabilities in other languages, particularly low-resource ones, are limited.\footnote{Code 
is available at \url{https://github.com/salesforce/FewXC}}
\end{abstract}

\section{Introduction}


It has long been known that NLP research and applications are concentrated on high-resource languages 
such as English, French, and Japanese. This limitation introduces bias and prevents people in minority language groups from accessing recent NLP technologies.

Driven by advances in large-scale training, there has been an increase in the number of approaches that attempt to learn general-purpose multilingual representations, which aim to capture shared knowledge across languages. Jointly trained multilingual language models such as XLM-R \cite{conneau-etal-2020-unsupervised} and mBART \cite{liu-etal-2020-multilingual-denoising}, coupled with supervised fine-tuning in the source (English) language, have been quite successful in transferring linguistic and task knowledge from one language to another without using any task labels in the target language, a.k.a. \emph{zero-shot transfer}. Despite their effectiveness, studies \citep{wu-dredze-2019-beto,pires-etal-2019-multilingual,k2020crosslingual} have also highlighted key factors for successful transfer which include structural similarity between languages and the tasks under consideration. When it comes to conversational tasks, studies on cross-lingual zero-shot transfer have been limited to only few domains and languages.



To investigate the cross-lingual transfer ability on conversational tasks, we create the XSGD dataset by translating data from the English-only Schema-Guided Dialogue or SGD~\citep{rastogi2020towards}, which is currently the largest multi-domain dialogue corpora. While previous work such as Multi$^2$WOZ~\citep{hung-etal-2022-multi2woz} has also tried to expand monolingual datasets into multiple languages, it is primarily a translation of development and test dialogues from the English-only MultiWOZ dataset~\citep{budzianowski-etal-2018-multiwoz,zang-etal-2020-multiwoz} into Arabic, Chinese, German, and Russian. In contrast, XSGD comprises 106 languages (including English), with roughly 330k utterances and 10 domains per language, as compared to the 7 domains and 29.5k utterances per language in Multi$^2$WOZ. 

Recently, several studies~\citep{li-liang-2021-prefix, lester-etal-2021-power, hambardzumyan-etal-2021-warp} have shown the potential of prompt tuning. In particular,~\citet{tu-etal-2022-prompt} observed that prompt tuning can achieve much better cross-lingual transfer than model fine-tuning across multiple XTREME tasks~\citep{pmlr-v119-hu20b} using significantly fewer parameters.
In this work, we propose an efficient prompt-tuning-based method that utilizes soft prompts to obtain stronger cross-lingually aligned representations on the XSGD dataset. The aligned prompts enable models to learn cross-lingual representations that can improve cross-lingual retrieval. Additionally, we compare the performance of vanilla and NLI-based formulations on intent classification task. 
The latter utilizes label descriptions or label names in conjunction with utterances for entailment prediction. We find that it exhibits stronger few-shot cross-lingual generalization capability for English-only tuning. Finally, our experimental results on intent classification and slot filling demonstrate consistent performance improvements with our learned aligned prompts, especially in few-shot settings.

Our contributions are summarized as follows:
\begin{itemize}
    \item We have constructed a large parallel multilingual conversation corpus comprising 106 languages. We are releasing this dataset to facilitate and foster further research on multilingual conversation tasks. 
    \item We have also introduced an efficient prompt-tuning-based approach for aligning sentence representations across multiple languages. 
    \item We explored two different task formulations in the context of cross-lingual settings. We found that the NLI-based formulation demonstrated much stronger cross-lingual ability than the vanilla one, especially in few-shot settings.
    \item Our experiments shows that the aligned prompt we proposed is effective for cross-lingual transfer, particularly in the few-shot setting, where we observe significant gains. Our study also showns the benefits of our approach, even when compared to large language models (LLMs) such as text-davinci-003 and ChatGPT.
\end{itemize}



\section{Background}

\subsection{Multilingual Models}
    
Pre-trained multilingual language models, such as mBERT \cite{devlin-etal-2019-bert}, XLM-R \cite{conneau-etal-2020-unsupervised}, and mBART \cite{liu-etal-2020-multilingual-denoising} have demonstrated remarkable zero-shot cross-lingual transfer ability across a range of NLP tasks \cite{pires-etal-2019-multilingual,wu-dredze-2019-beto}. Moreover, some prior work, such as \citet{Artetxe-tacl_a_00288,luo-etal-2021-veco,zhang-etal-2019-ernie}, has leveraged parallel data to further enhance the cross-lingual transfer ability of these models through fine-tuning the entire architecture. Our work mainly explore a similar direction for conversation tasks, but with a more efficient approach where only a small portion of parameters are fine-tuned.

\subsection{Cross-lingual Benchmarks}

To evaluate zero-shot cross-lingual transfer ability, it is a standard practice to fine-tune the models exclusively on English tasks and then evaluate them on non-English test sets. XTREME~\citep{pmlr-v119-hu20b} is a widely used benchmark in this regard, comprising four categories of tasks: sentence classification, structure prediction, question answering, and retrieval. For conversation tasks, the emerging benchmark is MASSIVE~\citep{fitzgerald2022massive}, which includes around 1 million utterances across a range of languages\footnote{Although this dataset does not contain any dialogue as our created dataset XSGD, it is of higher quality. As a result, we will be using it as a benchmark for downstream tasks.}. 

\subsection{Prompt Tuning}
Recently, prompt tuning, where only a small amount of additional parameters (i.e. prompts) is added and tuned, but the original model is kept frozen. Much fewer parameters or no parameters
are tuned and thus the training is a lot more efficient.
Several studies~\cite{li-liang-2021-prefix, lester-etal-2021-power, hambardzumyan-etal-2021-warp} have shown that prompt tuning looks promising on many NLU tasks. More recently, \citet{tu-etal-2022-prompt} observe that prompt tuning can achieve significantly better cross-lingual transfer than fine-tuning across several XTREME tasks \citep{pmlr-v119-hu20b}, despite only tuning 0.1\% to 0.3\% of the parameters compared with whole model fine-tuning.

\section{XSGD Dataset}

Prior work has focused on enhancing pre-trained language models (PLMs) for either 
deeper understanding of conversational contexts or improved cross-lingual generalization. For example, ~\citet{wu-etal-2020-tod} and ~\citet{vulic-etal-2021-convfit} have explored adapting general-purpose English PLMs \citep{devlin-etal-2019-bert,roberta} by applying conversation-specific training objectives on large-scale English conversational corpus. 


One of the main challenges to achieve cross-lingual conversational capability is the lack of paired multi-lingual conversational corpus. In this work, we take the initiative on this challenge and create a multi-lingual dataset~\textsc{XSGD} on top of the SGD dataset~\citep{rastogi2020towards}. To this end, we leverage Google Translate API \footnote{\url{https://cloud.google.com/translate}} and translate the original SGD dataset into 105 languages. It is a context-aware translation. Because of the limitations of the translation API, the maxim context is set to 100 utterances in a dialogue per API call. A complete list of the 105 languages can be found in Appendix \ref{appendix:A}. We follow the same train, development, and test splits as in the original SGD dataset. 

\paragraph{Human Evaluation} Our parallel dataset is the largest multilingual TOD corpus (330k per language), however, it inherits noise from the translation API. It is prohibitively expensive to do full-scale manual quality control because of its scale across 106 languages\footnote{It is an interesting direction to explore how to improve the quality of this public dataset via an economically efficient way in the future, for example,~\citet{majewska-etal-2023-cross}. }. 

\begin{table}[h!]
\centering
\scalebox{0.8}{
\begin{tabular}{c|cc}
 \toprule
 \multicolumn{1}{c}{\bf Languages}  & \multicolumn{2}{c}{\bf Human Evaluation }   \\  & Fluency  & Meaning   \\ 
 \midrule
Indonesian  &  99\% &  98\% \\
Swahili & 100\%  & 100\% \\
Khmer  & 94\%  &  99\%  \\
Urdu & 97\% & 100\% \\
Hawaiian &   95\% & 99\%  \\
Yoruba  & 98\% & 100\%  \\
  \hline
  \end{tabular}
}
\caption{Data quality results with Human evaluation.  }\label{table:xsgdHumanEval}
\end{table}

We conduct human evaluation on 100 randomly sampled examples with workers from Amazon Mechanical Turk (AMT) on 6 low-resource languages (Indonesian, Swahili, Urdu, Khmer, Hawaiian, Yoruba) with different scripts\footnote{Two languages (Hawaiian, Yoruba) are not even supported by backbone model XLM-R}. Each sample is a translation pair that are randomly selected consecutive turns within each dialogue. 
For quality control purpose, we set up a quiz to test Turkers's language skills. Each assignment is evaluated by three different Turkers. Turkers who passed the quiz are asked to evaluate the translation pairs based on 2 individual qualities (meaning and fluency): whether adequately expresses the meaning of English text, and whether the translated text is fluent.
We provide our evaluation template of Hawaiian language in Figure~\ref{fig:turker} of Appendix.  
As shown in Table~\ref{table:xsgdHumanEval}, we notice the high quality of our dataset. Surprisingly, at least 98\% have the same meaning of English text.\footnote{We hypothesize the conversation domain is easier to get high translation quality.}.

In the next section, we show an efficient transfer learning method to use this large scale dataset for alignment pretraining. Then we further tune the aligned model on clean data with gold-labels so that noise will hopefully have a minor effect on our final model. Our evaluation dataset is also a high quality multilingual dataset. 


\section{Method}
\label{sec:method}




\begin{figure}[!t]
    \centering
    \includegraphics[width=0.35\textwidth]{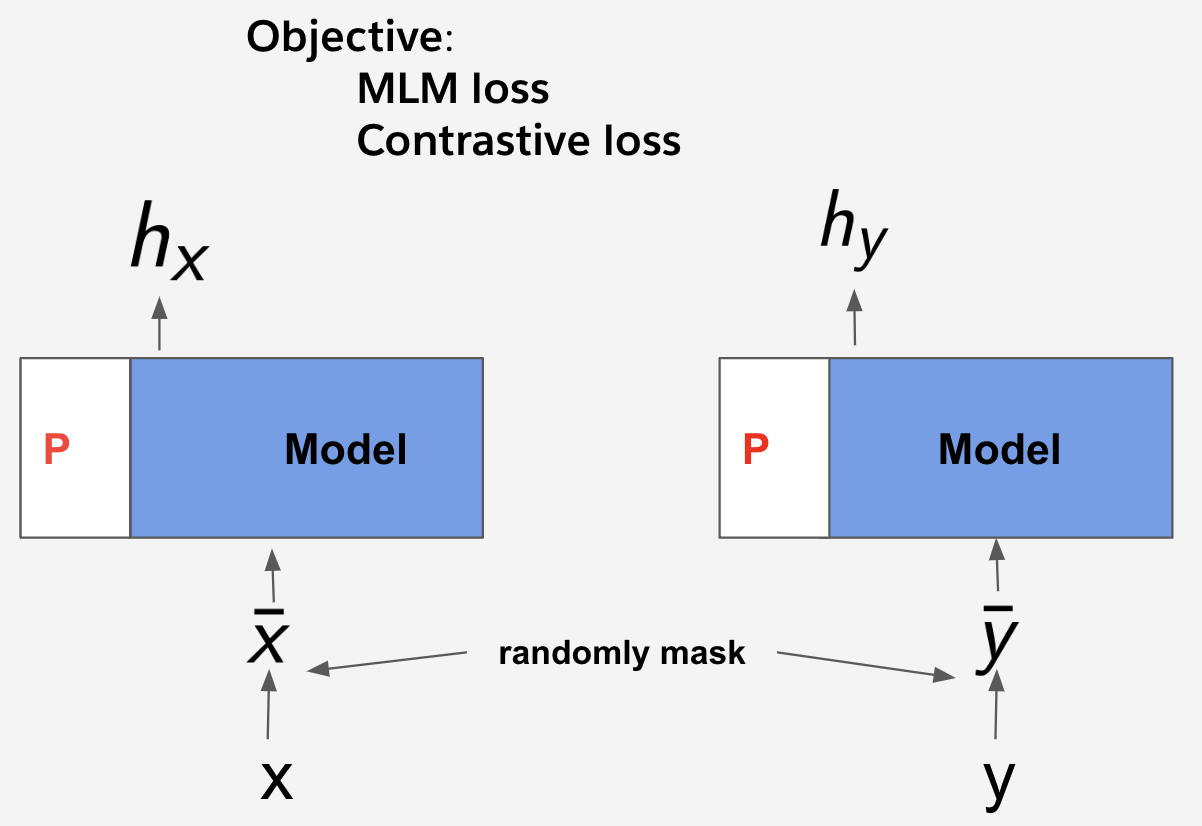}
    \caption{Framework for learning aligned prompts on multilingual conversational corpus. We denote \textcolor{red}{P} as the aligned prompts, which are tuned on the dialogue translation pairs, $\langle \x, \y\rangle$. The backbone model parameters are frozen. These aligned prompts are used for conversation downstream tasks.  }
    \label{fig:framework}
\end{figure}


In the zero-shot cross-lingual setting, models are fine-tuned solely on English and then evaluated on other languages. However, their performance on non-English languages, especially low-resource ones, tend to deteriorate \cite{pmlr-v119-hu20b,fitzgerald2022massive} 
. 

Previous works, specifically TOD-BERT~\citep{wu-etal-2020-tod}(with MLM loss) and ConvFiT~\citep{vulic-etal-2021-convfit} (with multiple negatives ranking loss), employ fine-tuning methods, where all model parameters are tuned. This process is not efficient for large pretrained models. The primary focus of our work is the exploration of efficient tuning methods.

To address this issue, we propose a prompt-tuning-based method that utilizes translation data to learn aligned prompts, which can lead to improved cross-lingual transfer performance, especially when task data in English is limited. 

\paragraph{Sequence Pairs} Our dialogue corpus consists of dialogues with approximately 20 turns each. To reduce the sequence length of each dialogue during training, we randomly select consecutive turns within each dialogue in each epoch and concatenate them into a sequence. We repeat this process for the corresponding turns in the target language.
We use this way to construct translation pairs dynamically during training, and then use the resulting translation pairs $\langle \x_i, \y_i\rangle$ from two different languages to learn aligned representations for an improved cross-lingual generalization capability\footnote{In our experiment, $x$ is always English.}. 

\paragraph{Masked Language Modeling (MLM) Loss} This is a popular learning objective to learn deep bidirectional representations. MLM is defined based on the reconstruction loss of a certain percentage of randomly masked input tokens given the rest of the context. 
We leverage this loss to adapt backbone models to the conversation domain. We conduct token masking dynamically during batch training. Formally, the MLM loss is defined as:
\begin{align}
\small
    &\mathrm{L}_{\textit{mlm}} = \nonumber \\ &-\frac{1}{M} \Bigg(\sum_{x_m \in \textit{MX}} \log \textit{prob}(x_m)+\sum_{y_m \in \textit{MY}}\log \textit{prob}({y_m})\Bigg)  \nonumber
\end{align}
where $M$ is the total number of masked tokens in $\langle \x, \y\rangle$ and $\textit{MX}$ and $\textit{MY}$ are the masked tokens in $x_i$ and $y_i$, respectively. $\textit{prob}(x_m)$ and $\textit{prob}(y_m)$ denote the probabilities of generating $x_m$ and $y_m$ from their corresponding masked tokens, respectively. 

In any pair of utterances $\langle \x, \y\rangle$, the dynamic mask strategy for $\x$ is independent of $\y$. During standard training, $\x$ is consistently set to English. However, $\langle \x, \y\rangle$ can represent any language pair among the 106 languages.

\paragraph{Contrastive Loss} 
We leverage contrastive learning to enhance the representations. And it would not be possible without our parallel data XSGD, which unlocks the possibility of learning stronger cross-lingual representations via alignment objective formulated via contrastive loss. Figure~\ref{fig:framework} illustrates the process. In a mini-batch of translation pairs, for $\langle \x, \y\rangle$, the positive sample for masked $x$ is the masked translation $y$. The negative samples are all the other translations $\hat{y}$ in the same mini-batch.

We first draw a batch of translation pairs. For each translation pair, we dynamically masked each sequence. The contrastive loss is
\begin{align}
\small
    \mathrm{L}_{\textit{contra}} = -\frac{1}{N} \Bigg(\sum_{\langle h_x, h_y\rangle \in H} \log \frac{\exp(\textit{sim} (h_x, h_y)/\tau)}{\sum_{y'} \exp( \textit{sim}(h_x, h_{y'})/\tau)}  \Bigg)  \nonumber
\end{align}
where $H$ is the translation representations of the batch,  $\tau$ is the temperature term, $N$ is the mini batch size, $\y'$ is from mini batch. $h_x$ and $h_y$ are the CLS token representations of masked sequence x and y respectively, $\textit{sim}$ is the similarity function. Cosine similarity is used in our experiments. We set $\tau=0.05$ in our experiments.

\paragraph{Total Loss} The overall learning objective is the sum of
 $\mathrm{L}_{\mathrm{mlm}}$ and $\mathrm{L}_{\mathrm{contra}}$.

\section{Experimental Setup}

\vspace{-0.2cm}
\begin{figure*}[h]
    \centering
    \vspace{-0.2cm}
    \includegraphics[width=0.7\textwidth]{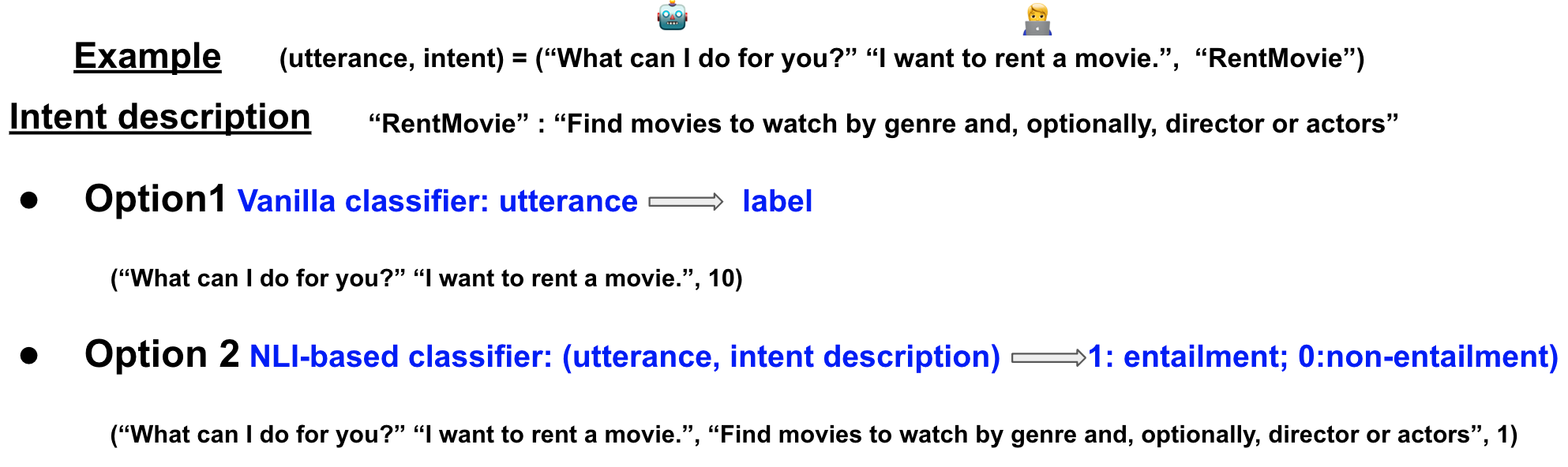}
    \caption{Two different classifiers (NLI-based classifier and vanilla classifier) are proposed for intent classification task. For NLI-based classifier training, negative samples are constructed in the mini batch. English intent description are also used for the evaluation on the other languages. See more details in \ref{sec:classifier}.}
    \label{fig:classifier}
\end{figure*}

\subsection{Datasets}
\paragraph{SGD} We use the Schema-Guided Dialogue (SGD) dataset~\citep{rastogi2020towards} for intent classification. There are about 16K dialogues and 20 domains.  For each domain, there are a different number of intents, services and dialogues. Each service provides a schema listing the supported intents along with
their natural language descriptions. For example, service ``payment'' have two intents ``MakePayment'' and ``RequestPayment''.  The description of an intent called ``MakePaymen'' is  ``Send money to your contact''.  Zero-shot evaluation is used, because lots of intents in the dev and test are unseen in the training set. For training, we only sample 5-shots per service as our training set and evaluate on the whole dev set. For cross-lingual evaluation, we use the translated utterance from XSGD\footnote{According to human evaluation results, we think it is reasonable to use them in some preliminary experiments. }.    

\paragraph{MASSIVE} We use MASSIVE~\citep{fitzgerald2022massive}
as another dataset for evaluation\footnote{We use the version MASSIVE 1.1, which can be downloaded at \url{https://github.com/alexa/massive}.}. There are 52 languages and about 1 million utterances in this dataset. For each language, there are about 11k train utterances, about 2k
dev utterances, about 3K test utterances. We use this for evaluation on two conversation understanding tasks: intent classification and slot filling. There are 60 intents and 55 slot types. Accuracy and F1 score are the metrics for intent classification and slot filling, respectively. 

\subsection{Task Classifiers}
\label{sec:classifier}

\paragraph{Intent Classifiers}
We use [CLS] representation from the encoder as the sentence representation. Two different intent classifiers (NLI-based classifier and vanilla classifier) are considered in our experiments. Figure~\ref{fig:classifier} shows more details. 

Vanilla classifier uses the utterance representation to predict intent label. The learning and inference is done as a multi-label classifier. 

NLI-based text classification has been investigated by \citep{qu-etal-2021-shot}, \citep{zhang-etal-2020-discriminative} and \citep{yin-etal-2019-benchmarking} and proved to show superior performance in few-shot setting. In NLI-based text classification scenario, utterance and intent description or intent name are combined to make a prediction. During training, positive samples are formed by concatenating utterance and its intent description. Negative samples are constructed in the mini batch by sampling a negative intent description. To balance the training process,  we keep the positive to negative ratio 1:1 for each batch. Cross-entropy loss is used during training. For inference, we select the label with largest entailment score. The prediction is correct if and only if the predicted label is correct and the largest entailment score is larger than 0.5 \footnote{The 0.5 threshold is for out-of-scope (OOS) prediction, which is required in the SGD dataset. The MASSIVE dataset doesn't have OOS, so the threshold can be disregarded.}.

\paragraph{Slot Classifier}
Slot filling is treated as a token level classification task. We report F1 score for this task on all languages.

\subsection{Training}
For the backbone model, we use XLM-R~\citep{conneau-etal-2020-unsupervised} in the most of experiments, which is a pretrained multilingual masked language model with 560M parameters on 2.5B of filtered data containing 100 languages. We also use XLM-RoBERTa-XL with 3.5B parameters in some settings. 
More details can be seen in Appendix~\ref{ap:training}. 



\section{Aligned Prompts Results}


In section~\ref{sec:method}, we propose a method that learns aligned prompts on conversation pair data in order to improve cross-lingual transfer ability. In this section, we show some aligned prompts results.

\begin{figure*}[h]
    \centering
    \includegraphics[width=0.6 \textwidth]{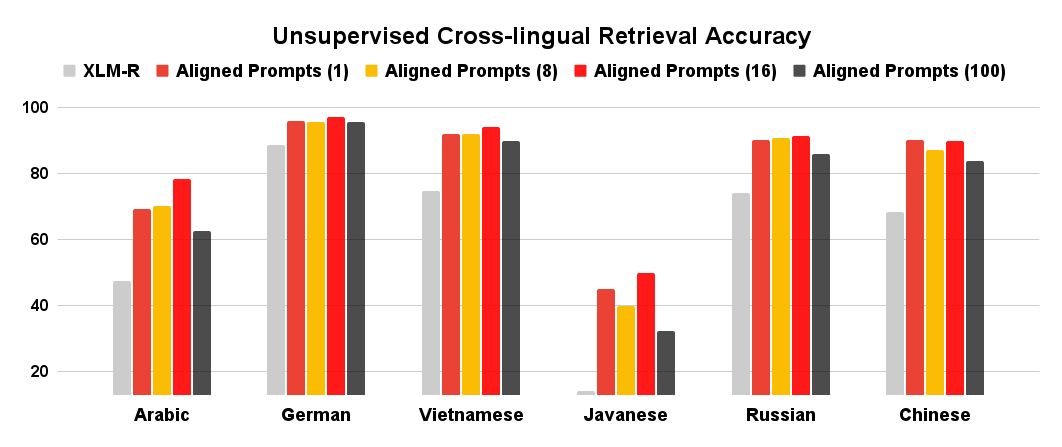}
    \caption{Unsupervised cross-lingual retrieval results (accuracy) for several linguistically diverse selected languages. The backbone model for these aligned prompts are XLM-R models. The length of prompts is 1, 8, 16, 100 respectively. XLM-R results are token from ~\citet{pmlr-v119-hu20b}.}
    \label{fig:retrivel-results}
\end{figure*}

\paragraph{Retrieval Results}
To justify what are the learn for these aligned prompts, we perform similarity search on Tatoeba, which is from from the XTREME benchmark~\citep{pmlr-v119-hu20b}. With aligned prompts, we use the CLS token representation as the sentence representation, and do nearest-neighbor search. Figure~\ref{fig:retrivel-results} displays the Tatoeba test results for several languages. Notably, our results demonstrate that aligned prompts can achieve significantly higher retrieval accuracy, even when the prompt length is only 1. Furthermore, performance can be further improved with additional prompts; however, it is important to note that using too many prompts can actually hurt performance. In our subsequent experiments, the prompt length was set to 16, unless otherwise specified. 

\vspace{-0.05cm}


\begin{table}[ht!]
\centering
\small
\begin{tabular}{c|c|c|c|}

 \multicolumn{1}{l|}{}  &  non-conversation & conversation  \\ 
\midrule
5-shots  & 51.7 (1.1) &  \textbf{55.2} (1.3)\\
\midrule 
 15-shots & 63.0 (0.5)  &  \textbf{66.5} (0.5) \\ 
 \midrule
 all-shots & 76.1 (0.6) &  \textbf{77.7} (0.5)  \\
\hline
\end{tabular}
\caption{Cross-lingual transfer (Training only on English annotation data, and evaluate on all languages) performance (with standard deviation) on intent classification when using aligned prompts from two different domains: conversation and non-conversation. All results are averaged over all languages of 5 runs. }
\label{table:domain}
\end{table}

\paragraph{Conversation Pairs vs. Non-Conversation Pairs}

Previous works have utilized parallel corpora from non-conversational domains, such as OPUS~\citep{tiedemann-2012-parallel}. To evaluate the effectiveness of XSGD, we randomly selected a parallel dataset from OPUS of a similar size and learned aligned prompts using the same method. Table~\ref{table:domain} presents the results of intent classification on a conversation downstream task, demonstrating that the performance of aligned prompts on XSGD significantly outperforms that of the non-conversational domain dataset across different settings (5-, 15-, all-shots).

\section{Downstream Tasks Results}
In this section, we perform experiments on a conversation benchmark MASSIVE and report the performance results on all languages. We try the following three tuning methods.

\paragraph{Fine-tuning (FT): } In this setting, all available parameters are tunable. 
\paragraph{Prompt Tuning (PT):} For prompt tuning, the backbone model is fixed, only a small number of parameters (prompts) and task classifiers parameters are updated. We use continuous prompts and layer prompts~\citep{li-liang-2021-prefix,liu-etal-2022-p}.

\paragraph{Aligned Prompt Tuning (APT):} With the parallel translation data, we can learn aligned prompt for aligned cross-lingual representation in Section~\ref{sec:method}. These prompts can be used for a warm-up start for these downstream task with prompt learning.



 


\begin{table}[ht!]
\centering
\small
\scalebox{0.9}{
\begin{tabular}{c|c|c|c|c|c|c|c|}

 \multicolumn{1}{l}{}  & en  & zh-CN & ja & ko & AVG   \\ 
\midrule
\multicolumn{4}{l}{\textbf{NLI-based Classifier}} \\
\midrule
5-shots & 47.8 &  31.3& 25.7 & 38.3  & \textbf{24.2} (6.8) \\ 
15-shots  & 70.8 & 53.1 &  43.5 & 61.8 & \textbf{46.0} (11.9)  \\
all  & 89.9  & 69.4 & 54.3 & 83.7 & 76.8 (0.6) \\
\midrule
\multicolumn{4}{l}{\textbf{Vanilla Classifier}} \\
\midrule
5-shots  & 9.4  & 4.4 &  4.2 &  6.6 &  5.9 (3.3) \\
15-shots  & 10.2 & 13.7 &  9.2 & 11.5 & 28.7 (17.3) \\
all  & 90.6  & 71.1   & 53.7  & 84.0  & \textbf{78.8} (0.5)  \\
 
\hline

\end{tabular}}
\caption{Averaged accuracy (\%)  of the NLI-based classifier and the vanilla classifier on the MASSIVE intent classification task when \textbf{fine-tuning} on English only and evaluating on all 52 languages. Results are averaged over all languages of 5 runs.
}
\label{table:intent_ft_result}
\end{table}

\subsection{Intent Classification}







\begin{table}[h!]
\centering
\small
\scalebox{0.9}{
\begin{tabular}{c|c|c|c|c|c|c|c|}

 \multicolumn{1}{l}{}  & en & zh-CN & ja & ko & AVG  \\ 

\multicolumn{4}{l}{\textbf{5-shots}} \\
FT  & 9.4    & 4.4 &  4.2 &  6.6 &  5.9 (3.3) \\
PT & 51.3    & 16.8 &  15.3 &  30.8& 24.9 (11.5) \\
APT  & \textbf{65.2}  & \textbf{52.1} &  \textbf{38.5} &  \textbf{59.3} & \textbf{55.2} (1.3)   \\
\midrule
\multicolumn{4}{l}{\textbf{15-shots}} \\
FT & 10.2  & 13.7 &  9.2 & 11.5 & 28.7 (17.4) \\
PT  & 75.8  & 56.5 &  43.6 & 63.7 & 58.2 (2.3) \\
APT  & \textbf{78.0} & \textbf{62.9} &  \textbf{47.7} & \textbf{71.7} & \textbf{66.5} (0.5)  \\
\midrule

\multicolumn{4}{l}{\textbf{all}} \\
\midrule

FT  & \textbf{90.6}  & \textbf{71.1}   & 53.7  & 84.0   & \textbf{78.8} (0.5)  \\
PT  & 89.7 & 68.2 & \textbf{55.6} & 82.1 & 76.8 (0.1) \\
APT  & 90.1  & 70.5  & 54.5  & \textbf{84.4}   & 77.7 (0.5) \\

\hline
\end{tabular}}
\caption{Accuracy (\%) of vanilla classifier on MASSIVE intent classification task when training on English only and evaluating on all 52 languages. Results are averaged over all languages of 5 runs.}
\label{table:van_result}
\end{table}



 

 


\begin{table}[h!]
\centering
\small
\scalebox{0.9}{
\begin{tabular}{c|c|c|c|c|c|c|}

 \multicolumn{1}{l}{}  & en & zh-CN & ja & ko & AVG   \\ 

\multicolumn{4}{l}{\textbf{5-shots}} \\
FT & 47.8 &  31.3& 25.7 & 38.3  & 24.2 (6.8)  \\
PT & 59.9    & 40.0 &  30.0 &  49.4&  38.1 (16.5) \\
APT  & \textbf{69.8}   & \textbf{52.4} &  \textbf{45.4} &  \textbf{64.8} & \textbf{59.8} (1.6)  \\
\midrule
\multicolumn{4}{l}{\textbf{15-shots}} \\
\midrule
 
FT  & 70.8 & 53.1 &  43.5 & 61.8 & 46.0 (11.9) \\
PT  & 75.8  & 57.8 &  43.5 & 68.7 & 60.3 (2.6)  \\
APT  & \textbf{89.7}  & \textbf{62.8} &  \textbf{51.8} & \textbf{75.0} & \textbf{67.5} (1.1)  \\

\midrule
\multicolumn{4}{l}{\textbf{all}} \\
FT  & 89.9 & \textbf{69.4} & \textbf{54.3} & 83.7 & 76.8 (0.6)  \\
PT  & 89.7  & 56.4 & 36.0 & 83.9 & 75.6 (0.4)  \\
APT  & \textbf{90.2}  & 68.4  & 52.0  & \textbf{85.2}   & \textbf{78.9} (0.2)  \\
 
\hline

\end{tabular}}
\caption{Accuracy (\%) of NLI-based classifier on MASSIVE intent classification task when training on English only and evaluating on all 52 languages. Results are averaged over all languages of 5 runs.}
\label{table:nli_result}
\end{table}

\paragraph{Fine Tuning} Table~\ref{table:intent_ft_result} shows the performance of the fine-tuned XLM-R model on English. Both of the intent classifiers achieve higher performance with more data. In few-shot experiments, the NLI-based classifier outperforms the vanilla classifier by a significant margin. The average performance on all 52 languages reaches 58.3\% accuracy with only 15 samples per intent. However, the vanilla classifier works better with the full data.

\paragraph{Vanilla Classifier} In Table~\ref{table:van_result}, we observe poor performance on few-shot settings for vanilla classifiers on intent tasks. However, significant gains are achieved with our method (from 5.9\% to 24.9\% on 5-shots and from 28.7\% to 58.2\% on 15-shots). We also observe that aligned prompts can further improve performance, with the best results obtained in few-shot settings. Additionally, the variances in task performance across all languages with aligned prompts are significantly smaller than fine-tuning and prompt tuning only. Although prompt tuning achieves higher accuracy on few-shot settings than fine-tuning, there is still a small gap, even with aligned prompts and full data training.

\paragraph{NLI-based Classifier} An advantage of using NLI-based classifiers is their ability to evaluate unseen intent labels if their descriptions are known. Additionally, we demonstrate strong performance on the SGD dataset. In Table~\ref{table:nli_result}, we present the results of fine-tuning with prompt tuning and aligned prompts for the MASSIVE dataset. With aligned prompts, we achieve strong accuracy results of 59.8\% on 5-shots and 67.7\% on 15-shots. Moreover, the English result on 15-shots with aligned prompts is comparable to the result obtained from full data training. These findings suggest that NLI-based classifiers with aligned prompts can efficiently learn with few samples. Aligned prompts consistently outperform other methods in this setting, indicating strong modeling ability and cross-lingual transfer ability.       

\paragraph{LLMs Results} We conducted experiments using both ChatGPT and the latest GPT-3.5 model (text-davinci-003 as of May, 2023) from OpenAI. We sampled 100 examples for each language and used the prompts provided in the Appendix. In the few-shot setting, the in-context examples were taken from the English partition. The intent classification results are presented in Table~\ref{table:llm_result}. The text-davinci-003 model showed significant improvements as more in-context examples were included, however, the ChatGPT model only demonstrated improvement in English. The cross-lingual ability of ChatGPT was found to be even worse, which led us to hypothesize that the data used to train ChatGPT is predominantly in English. Based upon these results, we can draw a conclusion that cross-lingual is still challenging in the era of LLMs, and smaller models still have an advantage over LLMs for the ability to quickly adapt into new domains through fine-tuning or prompt-tuning.

\begin{table}[h!]
\centering
\small
\scalebox{0.9}{
\begin{tabular}{c|c|c|c|c|c|}

 \multicolumn{1}{l}{}  & en  & AVG  \\ 

\multicolumn{1}{c}{\textbf{text-davinci-003}} \\
zero-shot &  59.0   & 40.8  \\
1-shot  &   71.0 &   51.2 \\
5-shot & 83.0 &  \textbf{54.6}\\
\midrule
\multicolumn{1}{c}{\textbf{ChatGPT}} \\
zero-shot  & 63.0 & \textbf{54.6}  \\
1-shot  & 76.0 & 51.2 \\
5-shots  &  \textbf{87.0} & 51.3 \\

\hline
\end{tabular}}
\caption{Accuracy (\%) of ChatGPT and text-davinci-003  on MASSIVE intent classification task.}
\label{table:llm_result}
\end{table}

\vspace{-0.2cm}

\paragraph{Takeaway} Upon analyzing the results presented in Tables~\ref{table:van_result} and~\ref{table:nli_result}, we can observe significant improvements with aligned prompts as compared to prompting tuning alone. For instance, the improvements for vanilla classifiers are 30.3\%, 8.3\%, and 0.9\% for 5-shots, 15-shots, and full data training, respectively. Similarly, for NLI-based classifiers, the gains are 11.7\%, 7.2\%, and 3.3\% for the same settings. We note that there is a clear trend where the gain of cross-lingual transfer ability decreases as more English training data is used. Furthermore, NLI-based classifiers exhibit superior cross-lingual transfer ability, particularly in the few-shot setting.      

\subsection{Slot Filling}

Table~\ref{table:slot_result} shows the evaluation results for slot filling using the XLM-R backbone model. Our models were trained solely on English data, but we report the results for all languages. However, the fine-tuned models' results for Chinese and Japanese are significantly worse than those for English. In fact, the gaps are much larger than those in a similar setting for the intent classification task. This observation suggests that slot filling is considerably more challenging than intent classification.

The performance differences between fine-tuning and prompt tuning for all languages averaged across are 6.4\%, -3.4\%, and -6.2\%, respectively. These results indicate that fine-tuning is more effective for improving slot filling performance than prompt tuning. However, this also suggests that there is still room for improvement for the current prompt-based methods.

With aligned prompts, we achieve consistent improvements over 5 runs, with gains of 4.5\%, 1.3\%, and 0.1\% in the averaged F1 score. These results are consistently better, but the improvements are smaller as the training dataset size increases.


\begin{table}[h!]
\centering
\small
\scalebox{0.9}{
\begin{tabular}{c|c|c|}
 \multicolumn{1}{l}{}  & en  & AVG   \\ 
\multicolumn{3}{l}{\textbf{5-shots}} \\
FT& 41.0 &   27.8 (3.3)   \\ 
PT&   59.5  &   34.2 (1.2)  \\ 
APT  & \textbf{62.6}   & \textbf{38.7} (0.9)  \\

\midrule
\multicolumn{3}{l}{\textbf{15-shots}} \\
\midrule

FT & 70.7 &  \textbf{49.0} (1.1)  \\
PT  & 70.9   & 45.6 (0.9) \\
APT  & \textbf{72.4}  & 46.9 (1.2)  \\
\midrule
\multicolumn{3}{l}{\textbf{all}} \\
FT  & \textbf{83.9}    & \textbf{61.6} (1.0)  \\
PT  & 83.3 & 55.4 (0.1) \\
APT  & 83.5  & 55.5 (0.5)  \\
\hline
\end{tabular}}
\caption{Slot filling F1 (\%) results on MASSIVE benchmark when training on English only and evaluate on all 52 languages.}
\label{table:slot_result}
\end{table}

\vspace{-0.3cm}




 

\paragraph{XLM-R-XL and OpenAI API Results} To test the limits of the prompt tuning method, we conducted experiments using prompt tuning and aligned prompts. Initially, we learned the aligned prompts on parallel XSGD data with a similar setting, where the prompt length is 16 and the backbone model is XLM-R-XL.

Table~\ref{table:slot_result} and Table~\ref{table:xl_result} displays the results of prompt tuning and aligned prompts on these settings. There are significant performance gains, particularly for aligned prompts. When scaling up the backbone model size from XLM-R to XLM-R-XL, the improvements with aligned prompts are 5.2\% and 5.0\% for 15-shots and full English data, respectively. Meanwhile, the improvements with prompt tuning are only 1.0\% and 0.5\%. This finding indicates that aligned prompts provide better modeling ability when increasing the backbone model size.

For the experiments with OpenAI models, we adapted prompts from ~\citet{Qin2023IsCA}. More details about the prompts and results are available in the Appendix. Overall, LLMs exhibit poor performance in the slot filling task, with an average F1 score ranging from 3\% to 6\% across all languages.

\begin{table}[ht!]
\centering
\small
\scalebox{0.9}{
\begin{tabular}{c|c|c|c|c|c|c|}

 \multicolumn{1}{l}{}  & en & zh-CN & ja  & AVG  \\ 

\midrule
\multicolumn{4}{l}{\textbf{15 shots}} \\
\midrule
PT &   71.7   & 10.1 &  5.1  &  46.6 (1.9) \\ 
APT  & \textbf{73.3}  & \textbf{22.1}  & \textbf{13.2}  &  \textbf{52.1} (0.5) \\
\midrule
\multicolumn{4}{l}{\textbf{all}} \\
\midrule
PT  & \textbf{83.1}   & 14.9 & 9.4  &   55.9 (0.7) \\
APT  & 82.8   & \textbf{23.6} & \textbf{11.7}  &  \textbf{60.5} (0.7)  \\
 
\hline

\end{tabular}}
\caption{Averaged Slot filling F1 (\%) results with 5 runs on MASSIVE benchmark when training on English only and evaluate on all 52 languages. The prompt lengths is 16. XLM-R-XL is used as the backbone model. }
\label{table:xl_result}
\end{table}

\paragraph{Discussion} We observe gains in cross-lingual ability with aligned prompts. 
However, there is still room for future improvements. The gains achieved with current aligned prompts methods are smaller than those achieved in few-shot settings. Also, the prompt tuning method on complex tasks, such as slot filling, still lags behind the fine-tuning method. These observations suggest that further research is needed to explore how to design more sophisticated and efficient methods for cross-lingual transfer.

\section{Related Work}

\paragraph{Methods for Cross-lingual Transfer}  

In recent years, many cross-lingual methods have been developed for non-conversational tasks using parallel data. However, continued pretraining on parallel data has been found to improve retrieval performance by making the pre-training task more similar to the downstream setting, but does not lead to a significant improvement in performance on other tasks~\citep{luo-etal-2021-veco,chi-etal-2021-improving,zhang-etal-2019-ernie}. These methods often require updating all model parameters or using larger scale monolingual corpora that cover all languages, which can make them difficult to use with large language models. In this work, we used a prompt-tuning-based method that only tunes few prompts and achieved significant gains in few-shot settings. We believe that more sophisticated work in this direction can be done in the future.

\paragraph{Resources for Multilingual Conversation} One of the fundamental objectives of artificial intelligence is to enable machines to communicate with humans. To achieve this, annotated conversation corpora are crucial. Conversation datasets have evolved from single-domain ones such as ATIS~\citep{price-1990-evaluation} to more complex and diverse ones such as MultiWOZ~\citep{budzianowski-etal-2018-multiwoz} and SGD~\citep{rastogi2020towards}. In recent years, several multilingual conversation datasets have been proposed to develop multilingual conversational models. However, most existing conversation systems are predominantly built for English or a few other major languages. For example, ~\citet{schuster-etal-2019-cross-lingual} introduced an annotation corpus of 57k utterances in English (43k), Spanish (8.6k), and Thai (5k) across three domains. Multi$^2$WOZ dataset~\citep{hung-etal-2022-multi2woz} is much larger annotation corpus with five languages (including English) and 29.5k utterances per language. Due to high cost for collecting multilingual conversation data, ~\citet{ding-etal-2022-globalwoz} introduces a novel data curation method for creating GlobalWoZ with 20 languages. In this work, we have created a new parallel multilingual dataset called XSGD by translating the English-only Schema-Guided Dialogue (SGD) dataset~\citep{rastogi2020towards} into 106 different languages. Although this dataset may contain some noise due to the translation process, we think it is a valuable resource for researchers interested in exploring multilingual conversational tasks.



\section{Conclusion}
In this paper, we present XSGD, a large-scale parallel multilingual conversation corpus that can be used for aligned cross-lingual transfer. Additionally, we propose a prompt-tuning method to learn alignment prompts, which can further improve the efficiency of the cross-lingual transfer. We evaluate our approach on intent classification and slot-filling tasks, and our experiments demonstrate its effectiveness. We also study popular LLMs and find that their performance on non-English languages remain to be improved.






\section*{Limitations}

Although the translated data can be a little noisy, in our work, we did not mainly use the data directly on downstream tasks. Instead, we propose an efficient transfer learning method to use this large scale dataset for alignment pretraining.  Then we further tune the aligned model on clean data with gold-labels so that noise will hopefully have a minor effect on our final model. Our evaluation dataset is also a high quality multilingual TOD dataset. So the proposed method and conclusion are still solid.


When conducting experiments with the OpenAI API, the large number of intent types (60) and slot types (55) posed a challenge in designing effective prompts. To address this, we conducted surveys and explored various prompt templates based on the works of ~\citet{Bang2023AMM,Qin2023IsCA,Lai2023ChatGPTBE}, among others. However, it is possible that we may have overlooked some potential prompt templates. There is room for improving the performance of text-davinci-003 and ChatGPT in future iterations.

We acknowledge that there are other parameter-efficient tuning techniques~\citep{pmlr-v97-houlsby19a,hu2022lora,ben-zaken-etal-2022-bitfit} and other LLMs, such as BLOOM~\citep{Scao2022BLOOMA1} and LLamA~\citep{Touvron2023LLaMAOA}. It is however nontrivial to compare against different parameter efficient methods on various different LLMs, which requires a significant amount of GPU hours and can warrant a paper by itself. Our contribution includes the massive XSGD multilingual data and an effective prompt-tuning based alignment method. We leave the exploration of other methods as future work.


\bibliography{anthology,custom}

\appendix
\section{Languages Except English on XSGD} \label{appendix:A}
List of 105 language ISO-639 code (\url{https://cloud.google.com/translate/docs/languages}) translated through Google Translate API (English is not included):
af, am, ar, az, be, bg, bn, bs, ca, ceb, co, cs, cy, da, de, el, eo, es, et, eu, fa, fi, fr, fy, ga, gd, gl, gu, ha, haw, he, hi, hmn, hr, ht, hu, hy, id, ig, is, it, ja, ka, kk, km, kn, ko, ku, ky, la, lb, lo, lt, lv, mg, mi, mk, ml, mn, mr, ms, mt, my, ne, nl, no, ny, or, pa, pl, pt, ro, ru, rw, si, sk, sl, sm, sn, so, sq, sr, st, su, sv, sw, ta, te, tg, th, tk, tl, tr, tt, ug, uk, ur, uz, vi, xh, yi, yo, zh-CN, zh-TW, zu

\section{Licenses of Datasets }
\begin{itemize}
    \item SGD~\citep{rastogi2020towards}: Attribution-ShareAlike 4.0 International Public License.
    \item Massive~\citep{fitzgerald2022massive}: Apache License.
    \item XSGD created by us: Attribution-ShareAlike 4.0 International.
\end{itemize}

\section{More Training Details}
\label{ap:training}
For the aligned prompts learning, we use Adam optimizer \citep{kingma2015adam} with warm up rate 0.1 and learning rate $\mathrm{e}\!-\!3$. The number of epoch is 10. The mini-batch size are 64 and 32 for XLM-R and XLM-RoBERTa-XL, respectively.   

On the conversation downstream tasks, we tune the learning rate in $\{0.1, 5\mathrm{e}\!-\!2, 2\mathrm{e}\!-\!2, 1\mathrm{e}\!-\!2, 5\mathrm{e}\!-\!3, 2\mathrm{e}\!-\!3, 1\mathrm{e}\!-\!3\}$.
For experiments on XSGD,  we do fine-tuning for 3 epochs and prompt-tuning for 30 epochs. For Massive benchmark, we fine tuning on intent classification and slot filling task for 30 epochs. For prompt tuning, the max number of epoch is 1000. We do early stopping based on performance on the English dev set. 1 A100 GPU with 40G memory is used for experiments. And most experiments are done in one day.

\section{Ablation Study on Learning Objectives}

An ablation study was conducted to analyze the learning losses for three different settings: prompt tuning (PT), aligned prompts (APT), and APT (with MLM only). The results on XSGD are shown in Figure~\ref{table:xsgd_result_ab}, while the results on MASSIVE intent classification can be seen in Figure~\ref{table:van_mlm_result}.

Please note that there is a comparison between MLM-only pre-training and MLM + Contrastive Loss on the parallel data:
\begin{itemize}
    \item APT (with MLM only): MLM-only pre-training
    \item APT: MLM + Contrastive Loss 
\end{itemize}

\begin{table}[ht!]
\centering
\small
\scalebox{0.9}{
\begin{tabular}{c|c|c|c|c|c|c|c|c|}

 \multicolumn{1}{l}{}  & en  &  hi & ms & vi & gd & tg & AVG \\ 

\midrule
\multicolumn{4}{l}{\textbf{Prompt Tuning}} \\
\midrule
l = 16  & 97.2 & 94.3 & 94.2 & 94.6 & 86.4 & 74.7 & 90.0 \\
\midrule
\multicolumn{4}{l}{\textbf{Aligned Prompts}} \\
\midrule
 & \textbf{97.7} &  \textbf{95.5} & \textbf{95.7} & \textbf{95.2} & \textbf{89.7} &  \textbf{75.3} & \textbf{91.4}\\

  \midrule
  \multicolumn{5}{l}{\textbf{Aligned Prompts (w/ MLM only)}} \\
 \midrule
  & 96.8 & 93.3 & 93.1 & 92.7 & 88.5 & 75.0 & 89.7 \\
\hline

\end{tabular}}
\caption{Intent classification accuracy (\%) on XSGD. Here we select some languages, which are in different language family or low-resourced. }
\label{table:xsgd_result_ab}
\end{table}

\begin{table}[h!]
\centering
\small
\scalebox{0.9}{
\begin{tabular}{c|c|c|c|c|}

 \multicolumn{1}{l}{}  & en  & AVG  \\ 

\multicolumn{3}{l}{\textbf{5-shots}} \\
PT & 51.3    &  24.9 (11.5) \\
APT  & \textbf{65.2}  & \textbf{55.2} (1.3)   \\
APT (w/ MLM only) & 61.9  & 30.9 (7.1) \\
\midrule
\multicolumn{3}{l}{\textbf{15-shots}} \\
PT  & 75.8 &  58.2 (2.3) \\
APT  & \textbf{78.0} & \textbf{66.5} (0.5)  \\
APT (w/ MLM only) &  78.2 & 61.2 (1.8) \\

\hline
\end{tabular}}
\caption{Accuracy (\%) of vanilla classifier on MASSIVE intent classification task when training on English only and evaluate on all 52 languages.}
\label{table:van_mlm_result}
\end{table}

\section{Prompt Templates and Results}
Prompt templates in experimental settings. \textbf{[schema]} and \textbf{[utt]} are the intent set and the raw utterance text respectively. And utt1, label1, utt2, label2 are in-context examples. 

\paragraph{Intent Classification Task}
\begin{quote}
    \textbf{Zero-shot Setting}\\
    \texttt{Please tell me the intent of the following utterance:[utt] given the intent set [schema]}
\end{quote}

\begin{quote}
    \textbf{Few-shots Setting}\\
    \texttt{Given the intent set [schema], please tell me the intent of the following utterances.\\\\utt1 \\ label1 \\ utt2 \\ label2 \\ ... \\ utt }
\end{quote}

\paragraph{Slot Filling Task}
\begin{quote}
    \texttt{Please identify slots {s} from the given text. The text from utt with slot annotations is formatted as [label : entity] .\\\\Text:[utt] \\Slot:
    }
\end{quote}


\begingroup
\begin{table*}[ht]
    \centering
    \vspace{2.8mm}
    \begin{tabular}{p{0.96\linewidth}}
        \toprule
    Please identify slots\ app\_name, currency\_name, radio\_name, email\_folder, relation, sport\_type, media\_type, music\_genre, drink\_type, ingredient, time\_zone, game\_name, weather\_descriptor, coffee\_type, podcast\_name, general\_frequency, transport\_type, time, playlist\_name, transport\_descriptor, movie\_name, cooking\_type, place\_name, device\_type, email\_address, change\_amount, timeofday, audiobook\_name, joke\_type, game\_type, transport\_agency, event\_name, song\_name, artist\_name, order\_type, person, player\_setting, house\_place, business\_name, food\_type, music\_album, meal\_type, definition\_word, podcast\_descriptor, transport\_name, audiobook\_author, date, movie\_type, music\_descriptor, list\_name, news\_topic, color\_type, Other, personal\_info, business\_type, alarm\_type from the given text. The text from utt with slot annotations is formatted as [label : entity]. \\ \\Text: \textcolor{green}{weck mich diese woche um fünf uhr morgens auf}\\Slot:\\\textcolor{blue}{app\_name : weck, currency\_name : None, radio\_name : None, email\_folder : None, relation : None, sport\_type : None, media\_type : None, music\_genre : None, drink\_type : None, ingredient : None, time\_zone : None, game\_name : None, weather\_descriptor : None, coffee\_type : None, podcast\_name : None, general\_frequency : None, transport\_type : None, time : fünf uhr morgens, playlist\_name : None, transport\_descriptor : None, movie\_name : None, cooking\_type : None, place\_name : None, device\_type : None, email\_address : None, change\_amount : None, timeofday : morgens, audiobook\_name : None, joke\_type : None, game\_type : None, transport\_agency : None, event\_name : None, song\_name : None, artist\_name : None, order\_type : None, person : None, player\_setting : None, house\_place : None, business\_name : None, food\_type : None, music\_album : None, meal\_ }
    \end{tabular}
    \caption{One example input and output pair for slot filling. The utterance and OpenAI API response are colored in \textcolor{green}{green} and \textcolor{blue}{blue}, respectively. }
\end{table*}
\endgroup

\begin{table*}[t!]
\centering
\scalebox{0.8}{
\begin{tabular}{ccccc|cc}
 \toprule
 \multicolumn{1}{c}{\bf Languages} & \multicolumn{4}{c|}{\bf Intent Classification } & \multicolumn{2}{c}{\bf Slot Filling }   \\
 & text-davinci-003 & ChatGPT & text-davinci-003 & ChatGPT  & text-davinci-003  & ChatGPT   \\ 
 & zero-shot & zero-shot & 5-shots & 5-shots  & zero-shot  & zero-shot   \\ 
 \midrule
   & Acc. & Acc. & Acc. & Acc.  & F1   & F1  \\ 
  \midrule 
 Afrikaans & 52 & 62 & 64 & 49 & 10.3 & 5.4  \\
Amharic & 5  & 14 & 13 & 8  & 0.0  & 0.0  \\
Arabic & 45 & 62 & 66 & 57 & 8.5  & 5.5  \\
Azerbaijani & 33 & 48 & 61 & 40 & 5.3  & 1.9  \\
Bengali & 32 & 56 & 45 & 46 & 3.0  & 1.9  \\
Catalan & 45 & 64 & 55 & 52 & 6.6  & 6.1  \\
Welsh & 21 & 31 & 34 & 21 & 2.9  & 2.0  \\
Danish & 62 & 70 & 72 & 65 & 12.7 & 5.3  \\
German & 55 & 76 & 76 & 72 & 13.6 & 5.4  \\
Greek & 45 & 66 & 67 & 75 & 7.9  & 3.7  \\
English & 59 & 63 & 83 & 87 & 23.8 & 1.6  \\
Spanish & 52 & 65 & 67 & 58 & 10.7 & 10.4 \\
Persian & 39 & 70 & 66 & 65 & 5.4  & 1.9  \\
Finnish & 45 & 62 & 62 & 49 & 5.3  & 3.5  \\
French & 54 & 78 & 77 & 73 & 12.9 & 8.8  \\
Hebrew & 42 & 64 & 60 & 55 & 1.6  & 0.0  \\
Hindi & 35 & 63 & 60 & 63 & 7.1  & 1.9  \\
Hungarian & 55 & 64 & 66 & 53 & 3.6  & 2.0  \\
Armenian & 11 & 26 & 21 & 22 & 0.0  & 5.5  \\
Indonesian & 55 & 60 & 70 & 63 & 11.1 & 1.9  \\
Icelandic & 46 & 57 & 49 & 40 & 4.7  & 3.6  \\
Italian & 60 & 66 & 67 & 63 & 6.0  & 5.3  \\
Japanese & 53 & 70 & 66 & 66 & 1.8  & 0.0  \\
Javanese & 19 & 15 & 25 & 21 & 1.6  & 0.0  \\
Georgian & 13 & 22 & 21 & 28 & 0.0  & 0.0  \\
Khmer & 15 & 22 & 34 & 18 & 4.3  & 2.0  \\
Kannada & 17 & 41 & 26 & 50 & 3.4  & 0.0  \\
Korean & 55 & 72 & 74 & 75 & 3.2  & 4.0  \\
Latvian & 41 & 49 & 52 & 41 & 1.7  & 7.2  \\
Malayalam & 17 & 40 & 27 & 40 & 1.6  & 5.6  \\
Mongolian & 14 & 24 & 30 & 25 & 0.0  & 0.0  \\
Malay & 51 & 49 & 66 & 55 & 11.7 & 1.9  \\
Burmese & 0  & 8  & 13 & 10 & 0.0  & 0.0  \\
Norwegian & 51 & 66 & 67 & 63 & 14.3 & 6.8  \\
Dutch & 63 & 71 & 71 & 64 & 12.8 & 5.8  \\
Polish & 60 & 64 & 71 & 68 & 13.2 & 1.8  \\
Portuguese & 53 & 62 & 65 & 60 & 14.5 & 10.5 \\
Romanian & 54 & 63 & 65 & 55 & 3.3  & 12.3 \\
Russian & 56 & 72 & 64 & 71 & 5.6  & 5.4  \\
Slovenian & 56 & 61 & 59 & 57 & 7.6  & 3.9  \\
Albanian & 39 & 41 & 47 & 35 & 6.2  & 2.0  \\
Swedish & 59 & 75 & 66 & 69 & 9.8  & 3.5  \\
Swahili & 21 & 47 & 27 & 34 & 0.0  & 3.6  \\
Tamil & 17 & 29 & 37 & 32 & 0.0  & 0.0  \\
Telugu & 22 & 33 & 32 & 31 & 0.0  & 0.0  \\
Thai & 50 & 62 & 69 & 69 & 3.5  & 4.0  \\
Tagalog & 49 & 58 & 59 & 51 & 10.1 & 6.2  \\
Turkish & 46 & 65 & 67 & 57 & 9.8  & 1.9  \\
Urdu & 18 & 52 & 30 & 46 & 3.5  & 2.0  \\
Vietnamese & 45 & 65 & 65 & 64 & 10.9 & 3.6  \\
Simplified Chinese & 60 & 75 & 74 & 64 & 0.0  & 0.0  \\
Traditional Chinese & 57 & 70 & 71 & 71 & 0.0  & 0.0 \\
  \hline
  AVG & 40.8 & 54.6 & 54.6 & 51.3 & & \\
  \end{tabular}
}
\caption{The performance results of the OpenAI API using our prompts on MASSIVE benchmark are presented. 100 examples are sampled for each language. For the slot filling task, the prompt used is adapted from ~\citet{Qin2023IsCA}. It should be noted that due to the large number of slot types (55), the slot results are not satisfactory. }\label{table:overview}
\end{table*}

\begin{table*}[t!]
\centering
\scalebox{0.8}{
\begin{tabular}{cc|c}
 \toprule
 \multicolumn{1}{c}{\bf Languages} & \multicolumn{1}{c|}{\bf Intent Classification } & \multicolumn{1}{c}{\bf Slot Filling }   \\
 & APT  & APT   \\ 
 & XLM-R  (NLI-based classifier) & XLM-R-XL   \\ 
 \midrule
   & Acc.  & F1  \\ 
  \midrule 
 Afrikaans & 78.5 &  66.5 \\
Amharic & 66.5  & 47.9 \\
Arabic & 72.8& 58.1 \\
Azerbaijani & 79.2& 61.7\\
Bengali & 80.3& 67.2 \\
Catalan &81.0 & 59.7\\
Welsh &62.6 & 52.1\\
Danish & 85.8  & 71.4\\
German & 84.2& 70.4 \\
Greek & 82.8 & 67.6\\
English & 90.1& 82.8 \\
Spanish & 84.2& 74.3\\
Persian & 85.9& 69.1\\
Finnish & 84.4&  73.1\\
French & 85.0&  65.1\\
Hebrew & 82.9& 49.4\\
Hindi & 83.9&  67.3\\
Hungarian & 82.5& 65.0\\
Armenian & 80.9& 60.5\\
Indonesian & 86.0& 67.2\\
Icelandic & 75.8& 60.3\\
Italian & 82.2& 67.8\\
Japanese & 55.6& 15.5\\
Javanese & 61.9&  46.8\\
Georgian & 72.0&  63.3\\
Khmer & 67.5&  53.3\\
Kannada & 76.8 & 62.2 \\
Korean & 86.0 & 65.8\\
Latvian & 80.6&  65.0 \\
Malayalam & 81.9& 66.7 \\
Mongolian & 79.4& 55.3 \\
Malay & 81.4& 66.3 \\
Burmese & 74.4& 59.2 \\
Norwegian & 85.5&  70.6\\

Dutch & 85.5& 70.6\\
Polish & 85.4& 65.5\\
Portuguese & 84.5& 67.0\\
Romanian & 83.4& 67.3\\
Russian & 85.3& 71.3 \\
Slovenian & 81.2& 67.0 \\
Albanian & 78.1& 58.7 \\
Swedish & 86.3& 75.9 \\
Swahili & 56.6& 43.7 \\
Tamil & 78.4& 60.3 \\
Telugu & 79.0& 65.1 \\
Thai & 81.7& 64.2 \\

Tagalog & 76.4 & 57.6 \\
Turkish & 82.3 & 64.6 \\
Urdu & 79.7 & 59.0 \\
Vietnamese & 83.8& 58.8 \\
Simplified Chinese & 69.3& 19.7 \\
Traditional Chinese & 67.3 & 19.2 \\
  \hline
AVG & 78.9 & 60.8 \\
  \end{tabular}
}
\caption{The performance results with Aligned Prompt Tuning (APT) on MASSIVE benchmark when training on English only and evaluating on all 52 languages. }\label{table:Ouroverview}
\end{table*}

\section{Amazon Mechanical Turk Template}
\begin{figure*}[h]
    \centering
    \includegraphics[width=0.9\textwidth]{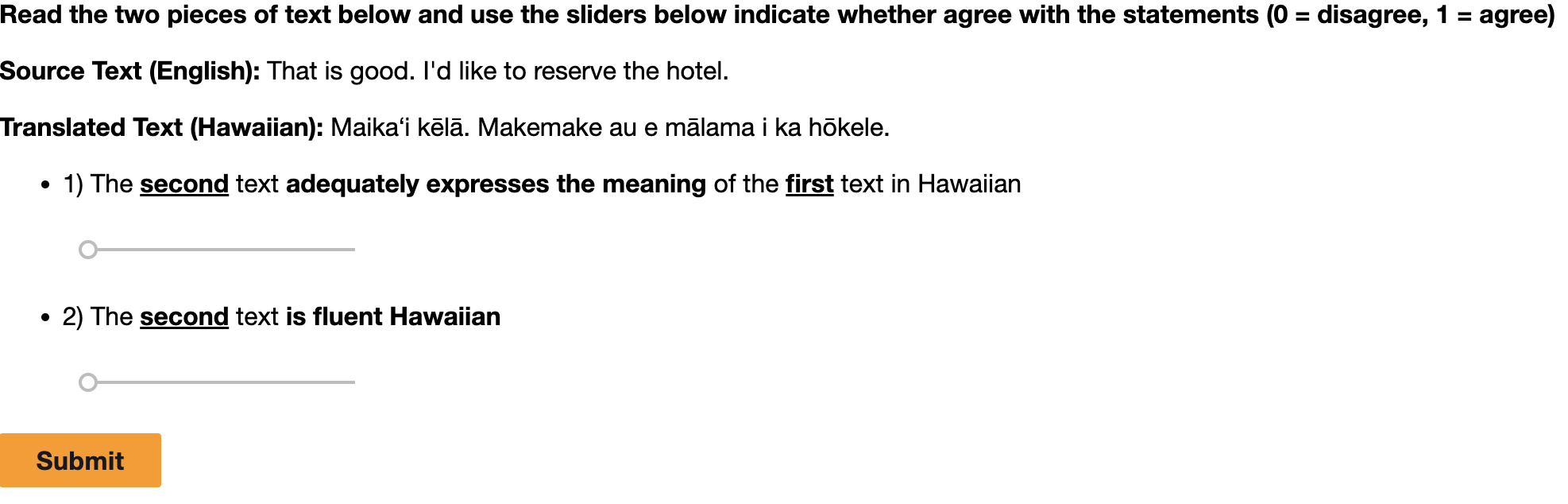}
    \caption{Human evaluation template for our dataset.}
    \label{fig:turker}
\end{figure*}

Please check one example in Figure~\ref{fig:turker} for human evaluation on XSGD.

\begin{figure*}[ht]
    \centering
    \includegraphics[width=0.8\textwidth]{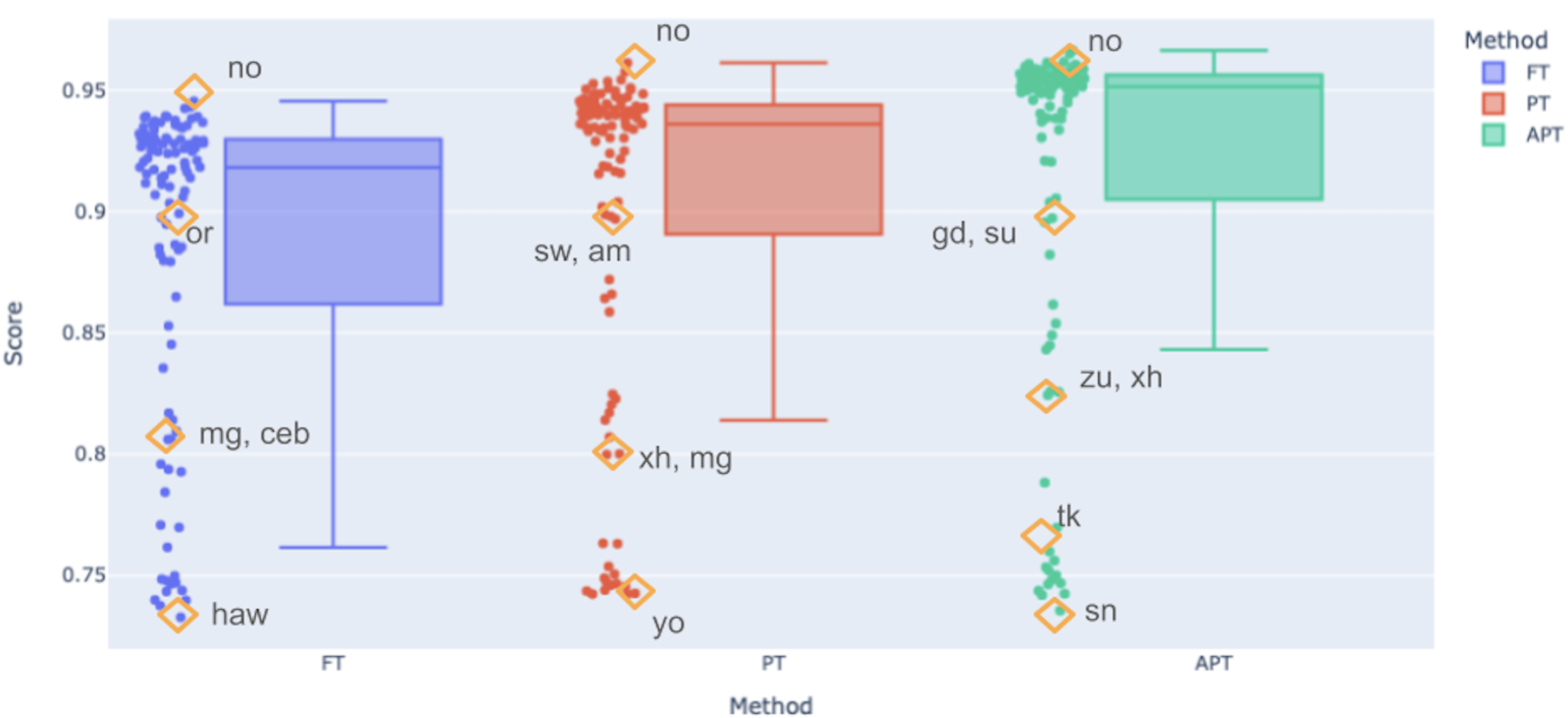}
    \caption{Intent classification performance of different models (FT: fine-tuning; PT: prompt tuning; APT: aligned prompt tuning) over all languages on XSGD. The scores represent the accuracy of each language. We can see the models are still struggled with languages that are not supported by the backbone model XLM-R.}
    \label{fig:xsgd}
\end{figure*}

\section{XSGD}

Table~\ref{table:xsgd_result} shows the intent classification results when training on English-only data and evaluating on all languages. We find that prompt tuning has better cross-lingual transfer ability and aligned prompts further improve the performance. 

\begin{table*}[ht!]
\centering
\small
\scalebox{0.9}{
\begin{tabular}{c|c|c|c|c|c|c|c|c|}

 \multicolumn{1}{l}{}  & en  &  hi & ms & vi & gd & tg & AVG \\ 
\midrule
\multicolumn{4}{l}{\textbf{Fine Tuning}} \\
\midrule
 & 95.7 & 92.8  & 93.2 & 93.9 & 84.5 & 75.0 & 88.6 \\ 

\midrule
\multicolumn{4}{l}{\textbf{Prompt Tuning}} \\
\midrule
l = 4  & 93.6  & 90.8  & 90.7  & 90.5 & 83.7 & 74.5 & 87.5\\
l = 8  & 96.2  & 94.4 & 93.8 & 94.7 & 85.8 & 74.3 & 89.8 \\
l = 16  & 97.2 & 94.3 & 94.2 & 94.6 & 86.4 & 74.7 & 90.0 \\
\midrule
\multicolumn{4}{l}{\textbf{Aligned Prompts}} \\
\midrule
 & \textbf{97.7} &  \textbf{95.5} & \textbf{95.7} & \textbf{95.2} & \textbf{89.7} &  \textbf{75.3} & \textbf{91.4}\\
 
\hline
\end{tabular}}
\caption{Intent classification accuracy (\%) on XSGD. Here we select some languages, which are in different language family or low-resourced. The monolingual training corpus size of “gd” for backbone model XLM-R is small ($\thicksim$0.1 GB). "tg" (Tajik) is also not supported by the backbone model. }
\label{table:xsgd_result}
\end{table*}

Figure~\ref{fig:xsgd} in the Appendix presents a performance comparison of the three different methods (FT: fine-tuning; PT: prompt tuning; APT: aligned prompt tuning). The figure indicates that prompt tuning outperforms fine-tuning, while aligned prompt tuning achieves the best performance. However, the models still struggle with some low-resource languages, especially those that are not supported by the backbone model XLM-R (e.g., haw (Hawaiian), yo (Yoruba), tk (Turkmen), sn (Shona)).

\end{document}